\documentclass{article}
\usepackage{ijcai21}

\usepackage{times}
\usepackage{soul}
\usepackage{url}
\usepackage[hidelinks]{hyperref}
\usepackage[utf8]{inputenc}
\usepackage[small]{caption}
\usepackage{subcaption}
\usepackage{graphicx}
\usepackage{amsmath, amssymb}
\usepackage{amsthm}
\usepackage{booktabs}
\usepackage{algorithm}
\usepackage{algorithmic}
\DeclareMathOperator{\E}{\mathbb{E}}

\title{Solving Continuous Control with Episodic Memory}

\author{
Igor Kuznetsov
\and
Andrey Filchenkov
\affiliations
ITMO University\\
\emails
igorkuznetsov14@gmail.com,
afilchenkov@itmo.ru,
}

\begin{document}

\maketitle

\begin{abstract}
  Episodic memory lets reinforcement learning algorithms remember and exploit promising experience from the past to improve agent performance. Previous works on memory mechanisms show benefits of using episodic-based data structures for discrete action problems in terms of sample-efficiency. The application of episodic memory for continuous control with a large action space is not trivial. Our study aims to answer the question: can episodic memory be used to improve agent's performance in continuous control? Our proposed algorithm combines episodic memory with Actor-Critic architecture by modifying critic's objective. We further improve performance by introducing episodic-based replay buffer prioritization. We evaluate our algorithm on OpenAI gym domains and show greater sample-efficiency compared with the state-of-the art model-free off-policy algorithms.
\end{abstract}

\section{Introduction}

The general idea of episodic memory in reinforcement learning setting is to leverage long-term memory reflecting data structure containing information of the past episodes to improve agent performance. The existing works~\cite{blundell2016model,pritzel2017neural}, show that episodic control (EC) can be beneficial for decision making process. In the context of discrete action problems, episodic memory stores information about states and the corresponding returns in a table-like data structure. With only several actions the proposed methods store past experiences in multiple memory buffers per each action. During the action selection process the estimate of taking each action is reconsidered with respect to the stored memories and may be corrected taking into the account past experience. The motivation of using episodic-like structures is to latch quickly to the rare but promising experiences that slow gradient-based models cannot reflect from a small number of samples.

The notion of using episodic memory in continuous control is not trivial. Since the action space may be high-dimensional, the methods that operate discrete action space become not suitable. 
Another challenge is the complexity of the state space. The study of ~\cite{blundell2016model} shows that some discrete action environments (e.g. Atari) may have high ratio of repeating states, which is not the case for complex continuous control environments.
    
 Our work proposes an algorithm that leverages episodic memory experience through modification of the Actor-Critic objective. Our algorithm is based on DDPG, the model-free actor-critic algorithm that operates over continuous control space. We offer to store the representation of action-state pairs in memory module to perform memory association not only from environment state, but also from the performed action. Our experiments show that modification of objective provides greater sample efficiency compared with off-policy algorithms. We further improve agent performance by introducing novel way of prioritized experience replay based on episodic experiences. The proposed algorithm, Episodic Memory Actor-Critic (EMAC), exploits episodic memory during the training, distilling the Monte-Carlo (MC) discounted return signal through the critic to actor and resulting in the strong policy that provides greater sample efficiency than other models. 

We draw the connection between the proposed objective and alleviation of Q-value overestimation, a common problem in Actor-Critic methods ~\cite{thrun1993issues}. We show that the use of the retrieved episodic data results in more realistic critic estimates that in turn provides faster training convergence. In contrast, the proposed prioritized experience replay, aims to focus more on optimistic promising experiences from the past. The process of frequently exploiting only the high-reward transitions from the whole replay buffer may result in unstable behaviour. However, the stabilizing effect of the new objective allows to decrease the priority of non-informative transitions with low return without loss in stability.

\begin{figure}[b]
     \centering
     \begin{subfigure}[b]{0.1\textwidth}
         \centering
         \includegraphics[width=\textwidth]{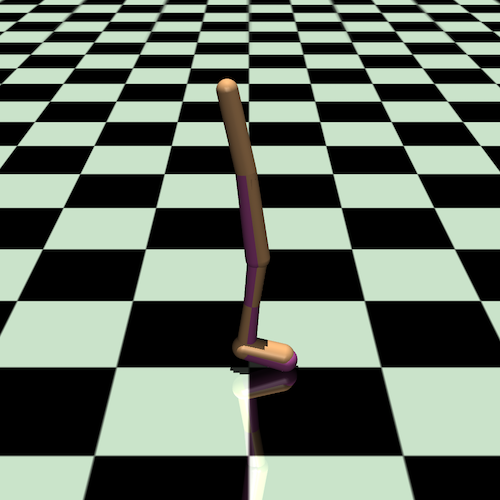}
         \caption{}
         \label{fig:walker}
     \end{subfigure}
     \begin{subfigure}[b]{0.1\textwidth}
         \centering
         \includegraphics[width=\textwidth]{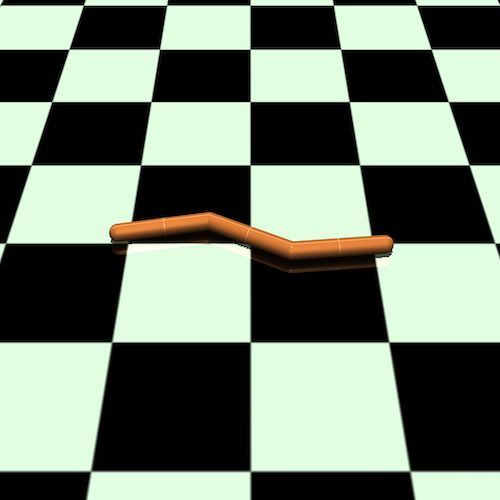}
         \caption{}
         \label{fig:swimmer}
     \end{subfigure}
     \begin{subfigure}[b]{0.1\textwidth}
         \centering
         \includegraphics[width=\textwidth]{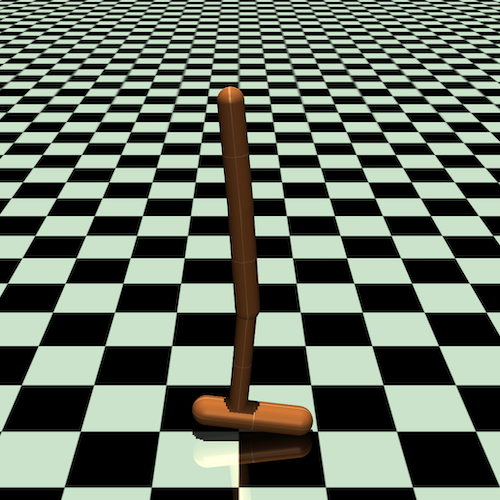}
         \caption{}
         \label{fig:hopper}
     \end{subfigure}
     \begin{subfigure}[b]{0.1\textwidth}
         \centering
         \includegraphics[width=\textwidth]{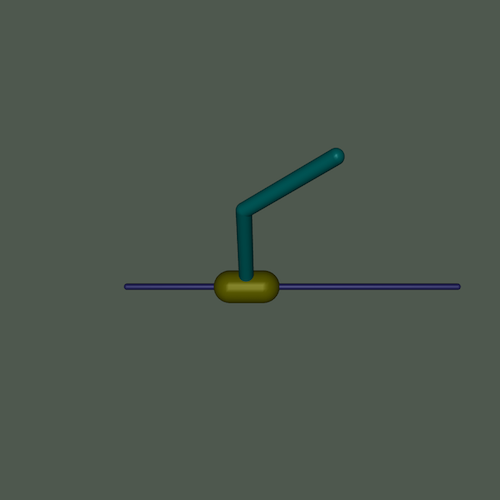}
         \caption{}
         \label{fig:inverted_double_pendulum}
     \end{subfigure}

        \caption{Example of OpenAI gym environments. (a) Walker2d-v3 (b) Swimmer-v3 (c) Hopper-v3 (d) InverteDoublePendulum-v2}
        \label{fig:envs}
\end{figure}

We evaluate our model on a set of OpenAI gym environments ~\cite{baselines} (Figure \ref{fig:envs}) and show that it achieves greater sample efficiency compared with the state-of-the-art off-policy algorithms (TD3, SAC).
We open sourced our algorithm to achieve reproducibility. All the codes and learning curves can be accessed at: \url{http://github.com/schatty/EMAC}.

\section{Related Work}

\paragraph{Episodic Memory.} First introduced in ~\cite{lengyel2007hippocampal}, episodic control is studied within the regular tree-structured MDP setup. Episodic control aims to memorize highly-rewarding experiences and replays sequences of actions that achieved high returns in the past. The successful application of episodic memory to the discrete action problems has been studied in ~\cite{blundell2016model} . Reflecting the model of hippocampal episodic control the work shows that the use of table-based structures containing state representation with the corresponding returns can improve model-free algorithms on environments with the high ratio of repeating states. Semi-tabular differential version of memory is proposed in ~\cite{pritzel2017neural} to store past experiences. The work of  ~\cite{lin2018episodic} is reminiscent of ideas presented in our study. They modify the DQN objective to mimic relationships between two learning systems of the human brain. The experiments show that this modifications improves sample efficiency for the arcade learning environment ~\cite{bellemare2013arcade}. The framework to associate the related experience trajectories in the memory to achieve reasoning of effective strategies is proposed in ~\cite{vikbladh2017episodic}. In the context of the continuous control, the work of ~\cite{zhang2019asynchronous} exploits episodic memory to redesign experience replay for the asynchronous DDPG. To the best of our knowledge, this is the only work that leverages episodic memory within the continuous control. 

\paragraph{Actor-Critic algorithms.} Actor-Critic methods represents a set of algorithms that compute value function for the policy (critic) and improve the policy (actor) from this value function. Using deep approximators for the value function and the actor, ~\cite{lillicrap2015continuous} presents Deep Deterministic Policy Gradient. The proposed model-free off-policy algorithm aims to learn policy in high-dimensional, continuous action space. Recent works propose several modifications that stabilize and improve performance of the original DDPG. The work of ~\cite{fujimoto2018addressing} addresses Q-value overestimation and proposes the Twin Delayed Deep Deterministic (TD3) algorithm which greatly improves DDPG learning speed. The proposed modifications include existence of multiple critic to reduce critic's over optimism, additional noise applied within calculation of target Q-estimate and delayed policy update. In ~\cite{haarnoja2018soft,haarnoja2018soft2} the authors study maximum-entropy objectives based on which provide state-of-the-art performance for OpenAI gym benchmark.

\section{Background}

Reinforcement learning setup consists of an agent that interacts with an environment \(E\). At each discrete time step \(t = 0...T\) the agent receives an environment state \(s_t\), performs an action \(a_t\) and receives a reward \(r_t\). An agent's action is defined by a policy, which maps the state to a probability distribution over the possible actions \(\pi: \mathcal{S} \rightarrow \mathcal{P} (\mathcal{A}) \). The return is defined as a discounted sum of rewards \(R_t=\sum_{i=t}^{T}\gamma^{i-t} r(s_i, a_i)\) with a discount factor \(\gamma \in [0, 1]\). 

Reinforcement learning aims to find the optimal policy \(\pi_\theta\), with parameters \(\theta\), which maximizes the expected return from the initial distribution \(J(\theta)= \E_{s_i \sim p_\pi, a_i \sim \pi } [R_0] \). The Q-function denotes the expected return when performing action \(a\) from the state \(s\) following the current policy \(\pi\) \(Q^\pi(s, a) = \E_{s_i \sim p_\pi , a_i \sim \pi} \left [ R_t | s, a \right] \)

For continuous control problems policy \(\pi_\theta\) can be updated taking the gradient of the expected return \( \nabla_\theta J(\theta) \) with deterministic policy gradient algorithm ~\cite{silver2014deterministic}:

\begin{align}
    \nabla_\theta J(\theta) = \E_{s \sim p_\pi } \left[ \nabla_{a} Q^\pi (s, a) |_{a=\pi(s)} \nabla_{\theta} \pi_\theta (s) \right].
\end{align}%

In actor-critic methods, we operate with two parametrized functions. An actor represents the policy \(\pi\) and a critic is the Q-function. The critic is updated with temporal difference learning by  iteratively minimizing the Bellman equation ~\cite{watkins1992q}:

\begin{align}
    J_Q = \E \left[ (Q(s_t, a_t) - (r(s_t, a_t) + \gamma Q(s_{t+1}, a_{t+1})))^2 \right].
\end{align}%

In deep Q-learning , the parameters of Q-function are updated with additional frozen target network \(Q_{\theta'}\) which is updated by \(\tau\) proportion to match the current Q-function \( \theta^{'} \leftarrow \tau \theta + (1-\tau) \theta^{'} \)

\begin{align}
    J_Q = \E \left[ (Q(s_t, a_t) - Q')^2 \right],
\end{align}%

where

\begin{align}
Q'=(r(s_t, a_t) + \gamma Q_{\theta'}(s_{t+1}, a')), a' \sim \pi_{\theta^{'}}(s_{t+1}).
\end{align}%

The actor is learned to maximize the current \(Q\) function. 

\begin{align}
    J_\pi = \E \left[ Q(s, \pi(s)) \right].
\end{align}%

The advantage of the actor-critic methods is that they can be applied off-policy, methods that proven to have better sample complexity ~\cite{lillicrap2015continuous}. During the training the actor and critic are updated with sampled mini-batches from the experience replay buffer ~\cite{lin1992self}. 

\section{Method}

We present Episodic Memory Actor-Critic, the algorithm that leverages episodic memory during the training. Our algorithm builds on the Deep Deterministic Policy Gradient ~\cite{lillicrap2015continuous} by modifying critic's objective and introducing episodic-based prioritized experience replay.

\subsection{Memory Module}

The core of our algorithm features a table-based structure that stores the experience of the past episodes. Following the approach of ~\cite{blundell2016model} we treat this memory table as a set of key-value pairs. The key is the representation of a concatenated state-action pair and the value is the true discounted Monte-Carlo return from an environment. We encode this state-action pair to a vector of smaller dimension by the projection operation \(\phi : x \rightarrow Mx, x \in \mathbb{R}^v, M \in \mathbb{R}^{u \times v} \), where \(v\) is the dimension of the concatenated state-action vector, and \(u\) is a smaller projected dimension. As the Johnson-Lindenstrauss lemma ~\cite{johnson1984extensions} states, this transformation preserves relative distances in the original space given that \(M\) is a standard Gaussian. The projection matrix is initialized once at the beginning of the training. 

Memory module implies two operations: add and lookup. As we can calculate the discounted return only at the end of an episode we perform add operation adding pairs of \(\phi([s, a])_i\), \(R_i\) when the whole episode is generated by the agent. We note that the true discounted return may be considered fair only when the episode ends naturally. For time step limit case we should perform additional evaluation steps without using these transitions for training. This way we allow discounted return from later transitions be obtained from the same amount of consequent steps as transitions in the beginning of the episode. Given the complexity and continuous nature of environments we make an assumption that the received state-action pair has no duplicates in the stored representations. As a result, we do not perform the search for the stored duplicates to make a replacement and we add the incoming key-value pair to the end of the module. Thus, the add operation takes \(O(1)\).

The lookup operation takes as input state-action pair, performs the projection to the smaller dimension and accomplishes the search for the \(k\) most similar keys returning the corresponding MC returns. We use \(l2\) distance as a metric of similarity between the projected states

\begin{align}
    d(z, z_i) = \left\Vert z - z_i \right\Vert^2_2 + \epsilon,
\end{align}%

where \(\epsilon \) is a small constant. The episodic MC return for a projected query vector \(z_q\) is formulated as a weighted sum of the closest \(K\) elements

\begin{align}
    Q_M(z_q) = \sum_{k=1..K} q_k w_k,
\end{align}%

where \(q\) is a value stored in the memory-module and \(w_i\) is a weight proportional to the inverse distance to the query vector.

\begin{align}
w_k = \frac{e^{-d(k,q)}}{\sum_{t=1..K} e^{-d(k,t)}}.
\end{align}%

We propose to leverage stored MC returns as the second pessimistic estimate of the Q-function. In contrast of approach of \cite{pritzel2017neural,lin2018episodic} we use weighted sum of all near returns rather then taking maximum of them. Given that we are able calculate MC return for each transition from the off-policy replay buffer we propose the following modification of the original critic objective (3):

\begin{align}
    J_Q =& (1-\alpha)(Q(s_t, a_t) - Q')^2 + \nonumber\\
    + & \alpha (Q(s_t a_t) - Q_{M})^2,
\end{align}%

where \(Q_M\) is the value returned by lookup operation and \(\alpha\) is a hyperparameter controlling the contribution of episodic memory. In the case of \(\alpha=0\) the objective becomes a common Q-learning loss function. In our experiments we found beneficial small values of \(\alpha\). In the evaluation results \(\alpha\) is set to 0.1 for the Walker, Hopper, InvertedPendulum, InvertedDoublePendulum, and to 0.15 for the Swimmer.

\begin{figure}[t]
    \centering
    \includegraphics[width=0.45\textwidth]{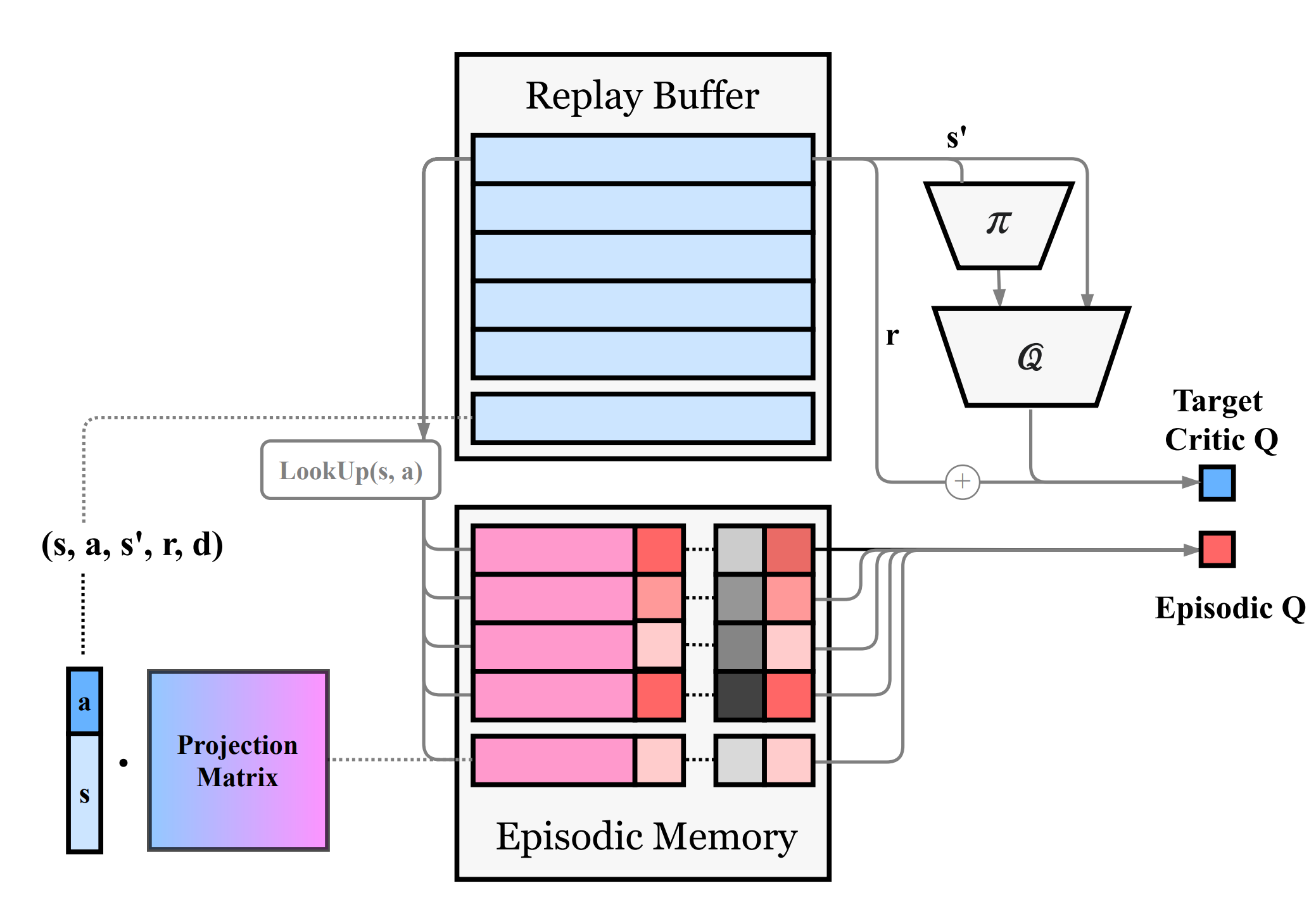}
    \caption{EMAC architecture on calculating Q-estimates}
    \label{fig:arch1}
\end{figure}

The memory module is similar to the replay buffer. They both needed to sample off-policy data during the Q-learning update. The difference is the memory module stores small-dimensional  representations of the state-action pair for effective lookup and discounted returns instead of rewards. The outline of the proposed architecture on calculating Q-estimates is presented in the Figure \ref{fig:arch1}.

\subsection{Alleviating Q-value Overestimation}

\begin{figure}
     \centering
     \begin{subfigure}[b]{0.214\textwidth}
         \centering
         \includegraphics[width=\textwidth]{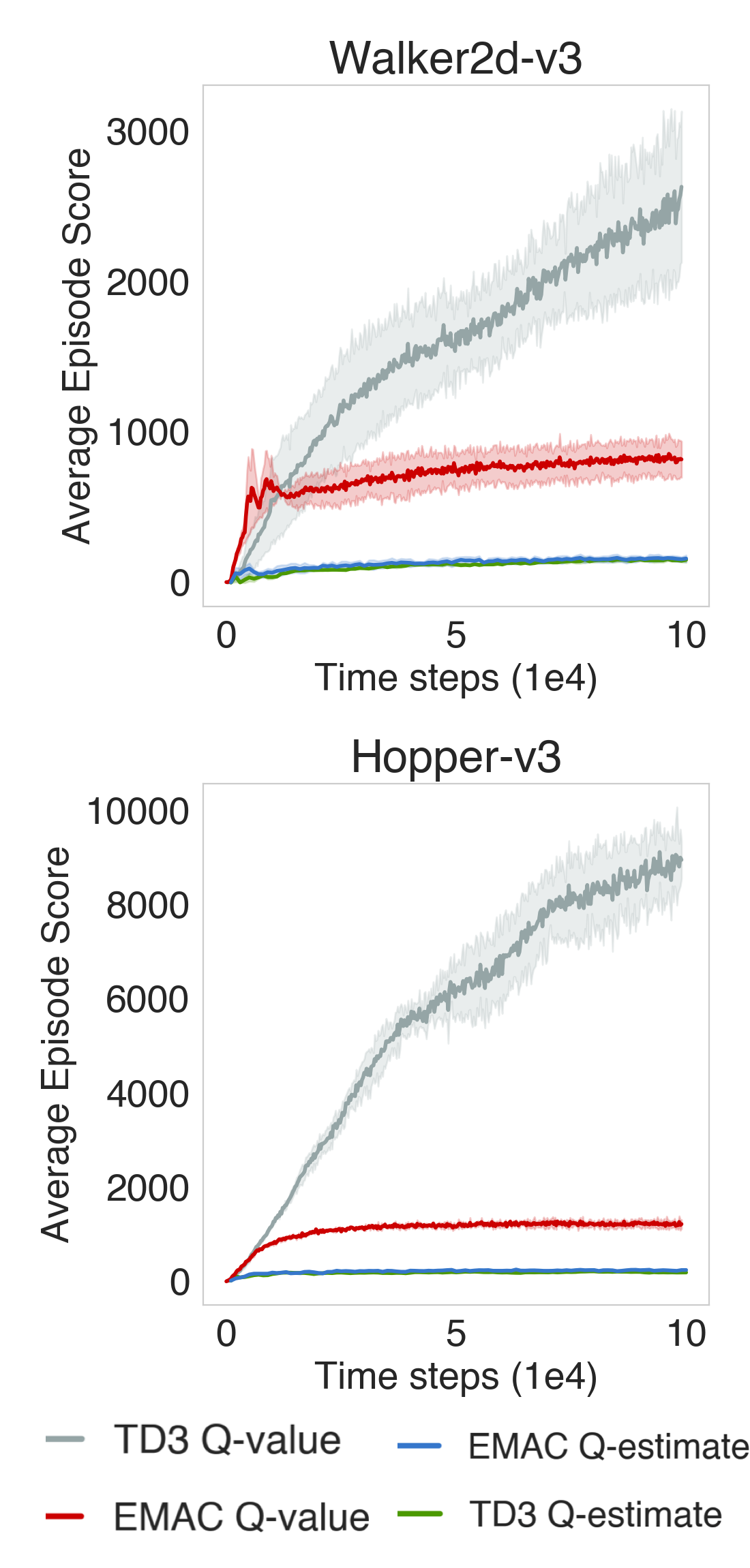}
         \caption{}
         \label{fig:q_over_a}
     \end{subfigure}
     \begin{subfigure}[b]{0.214\textwidth}
         \centering
         \includegraphics[width=\textwidth]{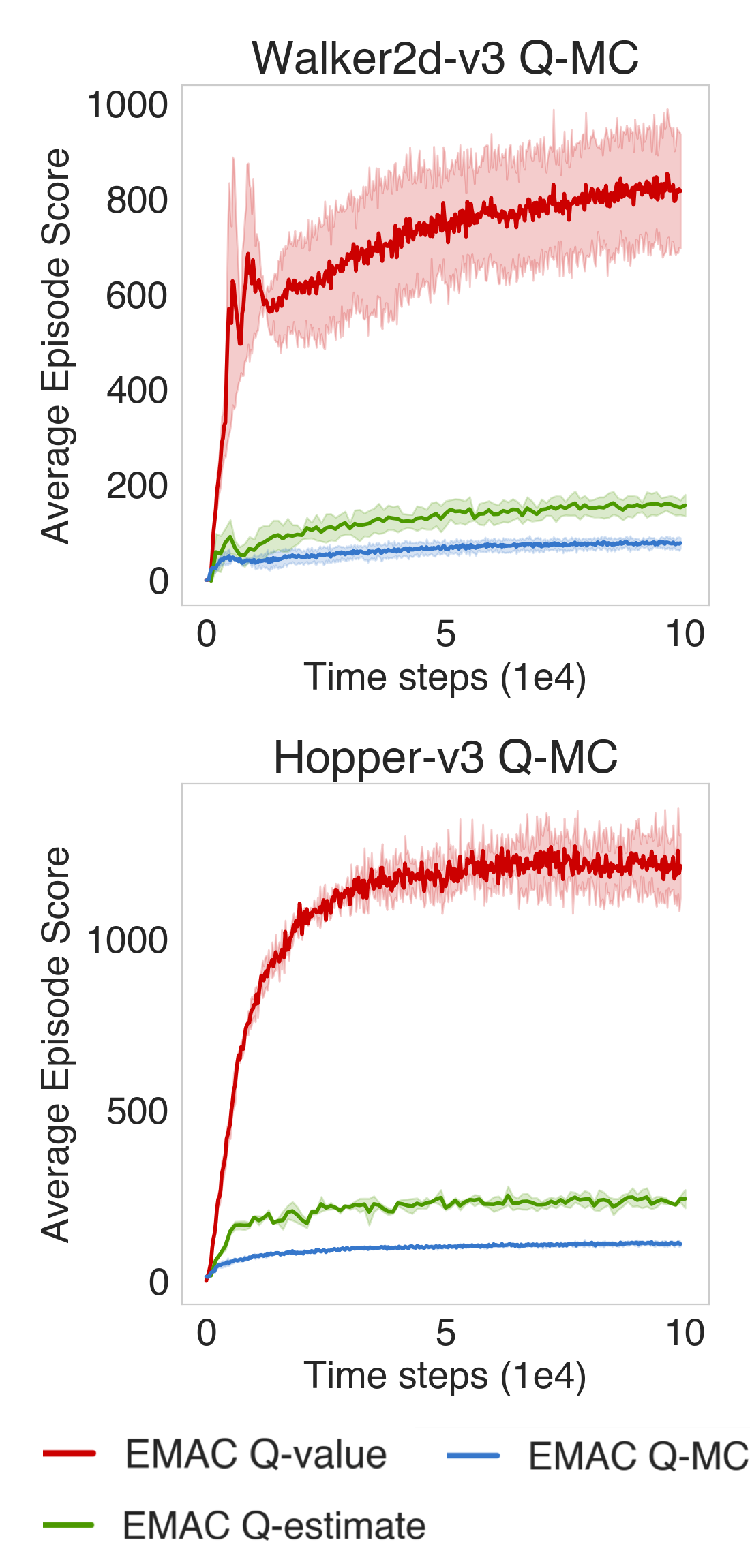}
         \caption{}
         \label{fig:q_over_b}
     \end{subfigure}

        \caption{Q-value overestimation. (a) Comparison between predicted Q-value and its true estimate. (b) Comparison between episodic stored MC estimate and true estimate}
        \label{fig:Q-value overestimation. }
\end{figure}

The issue of Q-value overestimation is a common case in Q-learning based methods ~\cite{thrun1993issues,lillicrap2015continuous}. Considering the discrete action setting, the value estimate is updated in a greedy fashion from suboptimal \(Q\) function containing some error \(y=r+\gamma \max_{a'} Q(s', a')\). As a result, the maximum over the actions along with its error will generally be greater than the true maximum ~\cite{thrun1993issues}. This bias is then propagated through the Bellman residual resulting in a consistent overestimation of the Q-function. The work of \cite{fujimoto2018addressing} studies the presence of the same overestimation in an actor-critic setting. The authors propose a clipped variant of Double Q-learning that reduces overestimation. In our work we show that additional episodic Q-estimate in critic loss can be used as a tool to reduce overestimation bias.

Following the procedure described in \cite{fujimoto2018addressing} we perform the experiment that measures the predicted Q-value from a critic compared with a true Q-estimate. We compare the value estimate of the critic, true value estimate and episodic MC return stored in the memory module. As a true value estimate we use a discounted return of the current policy starting from the state sampled randomly from the replay buffer. The discounted return is calculated from true rewards for maximum of 1000 steps or less in the case of the episode's end. We measure true value estimate each 5000 steps during the training of 100000 steps. We use the average of the batch for the critic's Q-value estimate and episodic return. The learning behaviour for the Walker2d-v3 and Hopper-v3 domains is shown in Figure \ref{fig:Q-value overestimation. }. In (a) we show that the problem of Q-value overestimation exists for both TD3 and EMAC algorithms, as both of Q-value predictions are higher than their true estimates. The Q-value prediction for EMAC shows less tendency for overestimaton. In (b) we compare true value estimate with episodic MC-return and show that latter has more realistic behaviour than the critic's estimate. Here, the difference between the true estimate and the MC-return is that episodic returns are obtained from the past suboptimal policy, whereas true value estimate is calculated using the current policy. The training curves from (b) show that MC-return has less value than the true estimate.

The experiment shows that episodic Monte Carlo returns from the past are always more pessimistic than the corresponding critic value. This fact makes incorporating MC-return into the objective beneficial for the training. As a result, the proposed objective with episodic MC-return shows less tendency to overestimation compared to the TD3 model. Given the state-action pair we here state that the MC return produced by the suboptimal policy for the same state may be used as a second stabilized estimate of the critic. Therefore, additional loss component of MSE between episodic return and critic estimate may be interpreted as a penalty for critic overestimation.

\begin{algorithm}[t]
\caption{EMAC}
\label{alg:algorithm}
\begin{algorithmic}[1] 
\STATE Initialize actor network \(\pi_\phi\), and critic network \(Q_\theta\)
\STATE Initialize target critic network \(Q_{\theta^{'}}\)
\STATE Initialize replay buffer  \(\mathcal{B}\), and episodic buffer \(\mathcal{B}_E\)
\STATE Initialize memory module \(\mathcal{M}\)
\FOR{t=1 to T}
\STATE Select action $a \sim \pi_\phi(s)$
\STATE Receive next state \(s^{'}\), reward \(r\), done \(d\)
\STATE Store \((s, a, r, s^{'}, d)\) in \(\mathcal{B}_E\)
\IF{\(d\)=True}
\FOR{i=1 to \(|\mathcal{B}_E|\)}
\STATE Calculate discounted episode return \(R_i\).
\STATE Store \((s_i, a_i, r_i, s^{'}_i, d_i)\) in \(\mathcal{B}\)
\STATE Add( \({[s_i, a_i], R_i}\) ) in \(\mathcal{M}\)
\ENDFOR
\STATE Free \(\mathcal{B}_E\)
\ENDIF
\STATE Sample mini-batch from \(\mathcal{B}\): \([s_b, a_b, r_b, s_b, r_b]\)
\STATE \(Q_M \leftarrow \) Lookup(\(s_b, a_b\)) from \(\mathcal{M}\)
\STATE Update critic \(Q_\theta\) by minimizing the loss:
\STATE $ J_{Q_\theta} = (1-\alpha)(Q - Q')^2 + \alpha (Q - Q_{M})^2 $
\STATE Update policy \(\pi_\phi\)
\STATE Update target critic \(\theta^{'} \leftarrow \tau \theta + (1-\tau) \theta^{'}\)
\ENDFOR
\STATE \textbf{return} \(\phi\)
\end{algorithmic}
\end{algorithm}

\subsection{Episodic-based Experience Replay Prioritization}

The prioritization of the sampled experiences for off-policy update of Q-function approximation is a studied tool for improving sample-efficiency and stability of an agent ~\cite{schaul2015prioritized,wang2016sample,kapturowski2018recurrent}. The common approach of using prioritization is formulated in the work of ~\cite{schaul2015prioritized}, where normalized TD-error is used as a criterion of transition importance during the sampling. We propose a slightly different prioritization scheme that is based on episodic return as a criterion for sampling preference. We formulate the probability of sampling a transition as

\begin{align}
    P(i) = \frac{p^\beta_i}{\sum_k p^{\beta}_k},
\end{align}%

where priority of transition \(p_i\) is a MC return stored in the memory module. The exponent \(\beta\) controls the measure of prioritization with \(\beta=0\) as a uniform non-prioritized case. The interpretation of such a prioritization is an increasing reuse of transitions which gave high returns in the past. The shift of true reward distribution used for off-policy update may result in divergent behaviour, therefore large values of \(\beta\) may destabilize the training. In our experiment the coefficient value of 0.5 gave promising results improving non-prioritized version of the algorithm.

\begin{figure*}[t]
    \centering
    \includegraphics[width=0.7\textwidth]{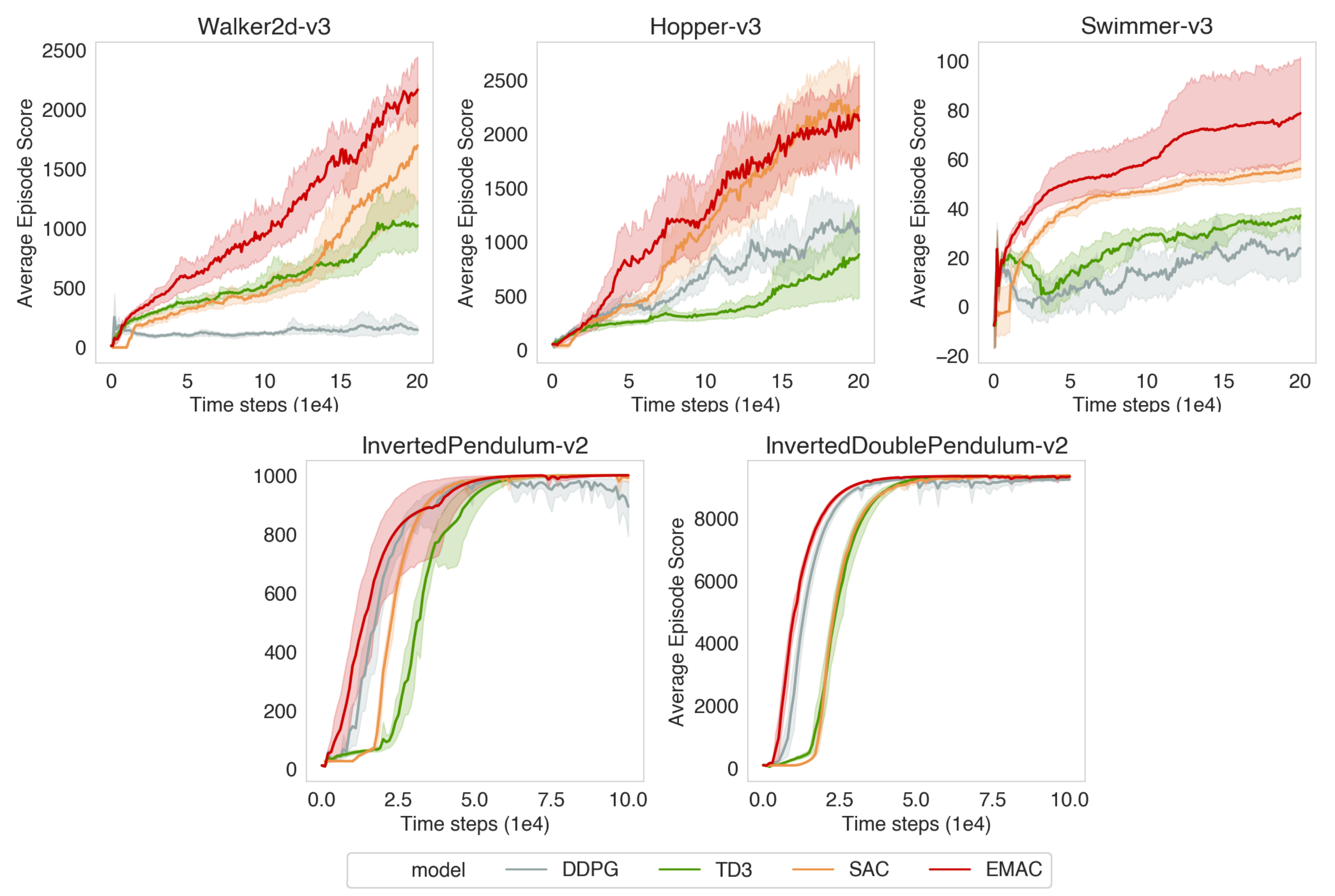}
    \caption{Evaluation results on OpenAI Gym Benchmark.}
    \label{fig:results_gym}
\end{figure*}

\begin{table*}
\centering
\begin{tabular}{lrrrrr}
\toprule
Environment                 &  EMAC                 & EMAC-NoPr             & DDPG     & TD3       & SAC    \\
\midrule
Walker2d-v3                 & $2236.88   \pm 808$     & $2073 \pm 768.88$       & 129.73   & 1008.32.  & 1787.28   \\
Hopper-v3                   & $1969.16   \pm 1142.58$ & $1837.3 \pm 794.17$     & 1043.78  & 989.96    & 2411.02   \\
Swimmer-v3                  & $80.53     \pm 33.07$   & $62.56  \pm 25.78$       & 27.45    & 38.91     & 56.7   \\
InvertedPendulum-v2         & 1000                  & 1000                  & 909.98.  & 1000      & 1000  \\
InvertedDoublePendulum-v2   & $9332.56   \pm 11.78$   & $9327.8 \pm 11.2$     & 9205.81  & 9309.73   & 9357.17        \\
\hline
\end{tabular}
\caption{Average return over 10 trials of 200000 time steps. \(\pm\) corresponds to a standard deviation over 10 trials.}
\label{tab:booktabs}
\end{table*}

The stored probabilities of sampling are recalculated after each episode comes to the memory-module and used consequently during sampling from off-policy replay buffer.

\section{Experiments}

The EMAC algorithm is shown in Algorithm \ref{alg:algorithm}. We compare our algorithm with model-free off-policy algorithms DDPG ~\cite{lillicrap2015continuous}, TD3 ~\cite{fujimoto2018addressing} and SAC ~\cite{haarnoja2018soft}. For DDPG and TD3 we use implementations of \cite{fujimoto2018addressing}. For SAC we implemented the model following the \cite{haarnoja2018soft}. All the networks have the same architecture in terms of the number of hidden layers, non-linearities and size of the hidden dimensions. In our experiments we focus on small-data regime given all algorithms 200000 time steps from an environment.

We evaluate our algorithm on a set of OpenAI gym domains ~\cite{baselines}. Each environment is run for \(200000\) time steps with the corresponding number of networks update steps. Evaluation is performed every 1000 steps with the reported value as an average from 10 evaluation episodes from different seeds without any exploration. We report results from both prioritized (EMAC) and non-prioritized (EMAC-NoPr) versions of the algorithm. The results are reported from 10 random seeds. The results in Table ~\ref{tab:booktabs} is the average return of the last training episode over all seeds. The training curves of the compared algorithms presented in Figure \ref{fig:results_gym}. Our algorithm outperforms DDPG and TD3 on all tested environments and SAC on three out of five environments.

Networks' parameters are updated with Adam optimizer ~\cite{kingma2014adam} with a learning rate of 0.001. All models consists of two hidden layers, size 256, for an actor and a critic and a rectified linear unit (ReLU) as a nonlinearity. For the first \(1000\) time steps we do not exploit an actor for action selection and choose the actions randomly for the exploration purpose.

\begin{figure}[t]
    \centering
    \includegraphics[width=0.45\textwidth]{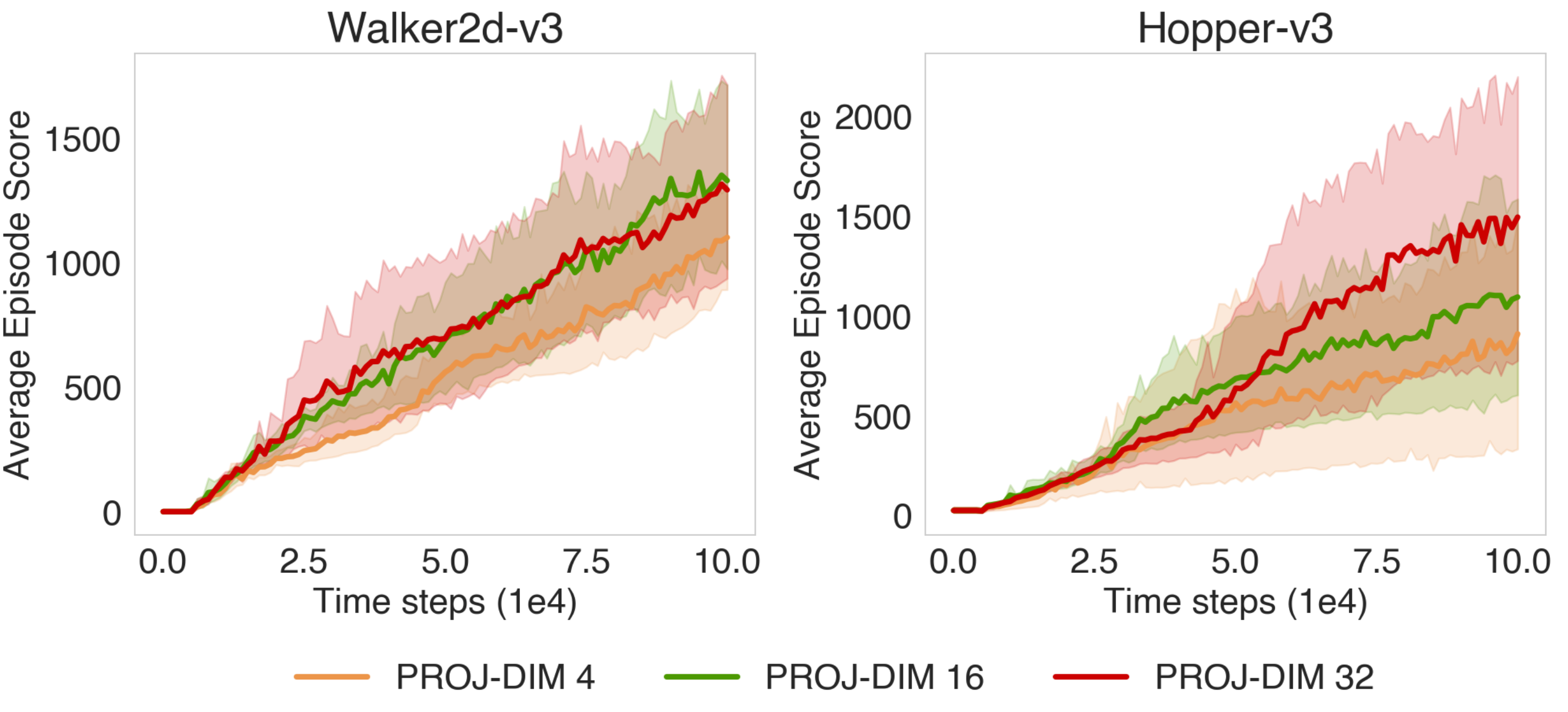}
    \caption{Performance of the EMAC with different size of projected state-action pairs}
    \label{fig:memory_dims}
\end{figure}

During the architecture design we study the dependency between the size of the projected state-action pairs and final metrics. We witness small increase in efficiency with greater projected dimensions.  Training curves showing different variants of projections of 4, 16, 32 sizes on domains of Walker-v3 and Hopper-v3 are depicted in Figure \ref{fig:memory_dims}. Unfortunately, increased projection size tends to slow down the lookup operation. In all our experiments we use the minimal option of projection size of \(4\) for the faster computations. The work of \cite{pritzel2017neural} leverages the KD-tree data structure to store and search experiences. On the contrary, we decided to keep table-based structure of the stored experiences, but placed the module on a CUDA device to perform vectorized l2-distance calculations. All our experiments are performed on single 1080ti NVIDIA card.

Due to low-data regime we are able to store all incoming transitions without replacement for the replay buffer as well as for episodic memory module. We set episodic memory size to 200000 transitions, although our experiments do not indicate loss in performance for cases of smaller sizes of 100 and 50 thousands of records. The \(k\) parameter that determines the number of top-k nearest neighbours for the weighted episodic return calculation is set to 1 or 2 dependent on the best achieved results from both options. We notice that k bigger than 3 leads to the worse performance.

\begin{figure}[t]
    \centering
    \includegraphics[width=0.43\textwidth]{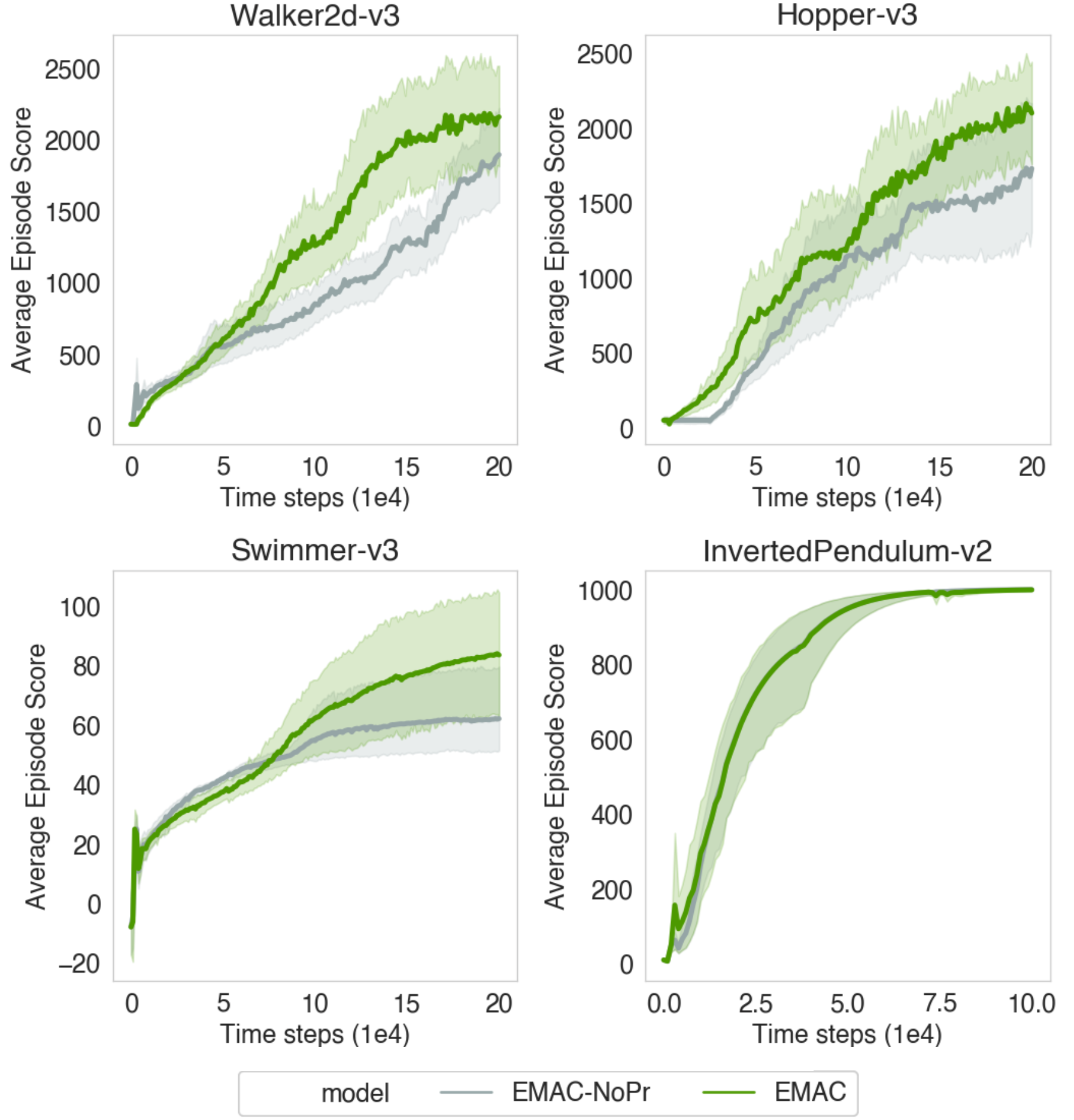}
    \caption{Comparison of prioritized (EMAC) and non-prioritized (EMAC-NoPr) versions}
    \label{fig:pior_study}
\end{figure}

To evaluate episodic-based prioritization we compare the prioritized and non-prioritized versions of EMAC on multiple environments. Prioritization parameter \(\beta\) is chosen to be 0.5. The learning behaviour of the prioritized and non-prioritized versions of the algorithm is showed in Figure \ref{fig:pior_study}. The average return for both versions is provided in Table \ref{tab:booktabs}.

\section{Discussion}

We present Episodic Memory Actor-Critic (EMAC), a deep reinforcement learning algorithm that exploits episodic memory in continuous control problems. EMAC uses non-parametric data structure to store and retrieve experiences from the past. The episodic memory module is involved in the training process via the additional term in the critic's loss. The motivation behind such an approach is that the actor is directly dependent on the critic, therefore improving critic's quality ensures the stronger and more efficient policy. We do not exploit episodic memory during the policy evaluation, which means that memory module is used only within the network update step. The loss modification demonstrates sample-efficiency improvement over the DDPG baseline. We further show that introducing prioritization based on the episodic memories improves our results. Experimental study of Q-value overestimation shows that proposed approach has less tendency in critic overestimation thus providing faster and more stable training.

Our experiments show that leveraging episodic memory gave superior results in comparison to the baseline algorithm DDPG and TD3 on all tested environments and also outperformed SAC on 3 out of 5 environments. We hypothesize that the applicability of the proposed method is dependent from environment complexity. As a result, we struggle to outperform SAC for such a complicated environment as Humanoid  which has bigger action space than the other OpenAI gym domains.

We now outline the directions of future work. As noted in ~\cite{daw2005uncertainty}, both animal and human brain exploits multiple memory systems while solving various tasks. Episodic memory is often associated with hippocampal involvement \cite{blundell2016model,lin2018episodic} as a long-term memory. Although the gradient-update-based policy may be seen as a working memory, it may be beneficial to study the role of separate short-term memory mechanisms alongside episodic memory for better decision making. Another interesting direction is to use different state-action representation for storing experiences. Although the random projection provides a way to transfer the distance relation from the original space to the smaller one, it does not show topological similarity between the state-action records. One possible way to overcome this issue is to use differential embedding for state-action representation. Unfortunately, the changing nature of embeddings entails the need of the constant or periodical memory update, which is alone an engineering challenge. We believe that our work provides the benefit of using episodic-memory for continuous control tasks and opens further research directions in this area.

\bibliographystyle{named}
\bibliography{ijcai21}

\end{document}